\documentclass[conference]{IEEEtran}
\IEEEoverridecommandlockouts
\usepackage{cite}
\usepackage{amsmath,amssymb,amsfonts}
\usepackage{algorithmic}
\usepackage{graphicx}
\usepackage{textcomp}
\usepackage{xcolor}
\usepackage{booktabs}
\usepackage{multirow}
\usepackage{url}

\def\BibTeX{{\rm B\kern-.05em{\sc i\kern-.025em b}\kern-.08em
    T\kern-.1667em\lower.7ex\hbox{E}\kern-.125emX}}
\begin{document}

\title{Prompt Privilege in Large Language Models: Measuring and Mitigating Barriers to Equitable AI Access} 

\author{

\IEEEauthorblockN{
\begin{tabular}{c c c}
\begin{tabular}{c}
Lier Jin\textsuperscript{$\dagger*$}\\
\textit{Fuqua School of Business}\\ 
\textit{Duke University}\\
Durham, USA\\
lierjin@alumni.duke.edu
\end{tabular}
& 
\begin{tabular}{c}
Lan Hu\textsuperscript{$\dagger$}\\
\textit{Department of Engineering}\\
\textit{Carnegie Mellon University}\\
Pittsburgh, USA\\
lanh@alumni.cmu.edu
\end{tabular}
&
\begin{tabular}{c}
Binqi Shen\\
\textit{Department of Industrial Engineering}\\
\textit{and Management Sciences}\\
\textit{Northwestern University}\\
Evanston, USA\\
binqishen2021@u.northwestern.edu
\end{tabular}
\end{tabular}
}

\vspace{1.5em}

\IEEEauthorblockN{
\begin{tabular}{c c}
\begin{tabular}{c}
Hanyu Cai\\
\textit{Department of Industrial Engineering}\\
\textit{and Management Sciences}\\
\textit{Northwestern University}\\
Evanston, USA\\
hanyucai2022@u.northwestern.edu
\end{tabular}
&
\begin{tabular}{c}
Yuting Xin\\
\textit{Department of Information}\\
\textit{and Decision Sciences}\\
\textit{University of Minnesota}\\
Minneapolis, USA\\
yuting.xin@outlook.com
\end{tabular}
\end{tabular}
}

\vspace{1.5em}

\IEEEauthorblockA{
\vspace{0.8em}
\textsuperscript{$\dagger$}Equal contribution \quad
\textsuperscript{*}Corresponding author
}
}

\maketitle

\begin{abstract} Large language models (LLMs) are increasingly used in healthcare, education, public services, and everyday decision making. They should provide comparable assistance regardless of users’ literacy, communication style, or prompt-engineering expertise. However, research on prompt robustness primarily focuses on adversarial attacks, prompt injection, and prompt optimization, overlooking whether semantically equivalent requests receive different responses simply because they are phrased differently. We call this accessibility challenge “Prompt Privilege”: users with greater prompting expertise obtain better model performance despite expressing the same underlying intent.

To address this problem, we propose the Prompt Equity Score (PES), a metric for evaluating performance consistency across user populations, and the Prompt Equity Transformer (PET), an LLM-based agent that automatically transforms user requests into semantically equivalent prompts while preserving their intent. PET shifts prompt optimization from users to AI systems, serving as an accessibility layer between users and foundation models. Experiments on MedQA reveal measurable prompt privilege, with statistically significant performance disparities between low-literacy and expert-prompting cohorts. Applying PET reduces these differences, with no pairwise cohort differences reaching statistical significance in our evaluation. These results suggest that accessibility-oriented prompt normalization can improve equitable AI access. By introducing prompt privilege as a new dimension of AI accessibility and PET as a practical solution, this work advances system-centered accessibility and provides a foundation for fairer, more trustworthy, and more inclusive AI systems.

\end{abstract}


\section{Introduction}
Artificial intelligence (AI) systems are increasingly deployed as decision-support tools across domains such as healthcare \cite{chu2026ct,zhou2025adhera}, financial decision intelligence \cite{han2026interpretable}, and autonomous driving \cite{zhou2023fastpillars,zhou2024lidarptq}. Trustworthy interaction design has therefore emerged as an important research direction as conversational AI becomes increasingly integrated into human-centered applications \cite{zhou2025beyond,zhou2026risk,zhang2026performance}. As these systems become more widely adopted, their usefulness depends not only on their underlying capabilities but also on whether users from diverse backgrounds can obtain comparable performance despite communicating differently. Ideally, AI systems should provide consistent assistance regardless of a user's health literacy, digital literacy, language proficiency, or familiarity with prompt engineering. In practice, however, users communicate with LLMs using substantially different linguistic styles and varying levels of technical sophistication, raising an important question about whether current systems provide equitable access to their capabilities.

A substantial body of work demonstrates that LLM performance is highly sensitive to prompt formulation. Chain-of-Thought prompting \cite{wei2022chain} and Automatic Prompt Engineer (APE) \cite{zhou2023large} show that carefully designed prompts can substantially improve reasoning performance without modifying model parameters, while work on adversarial prompt robustness \cite{zhu2023promptbench} and on how demonstrations and prompt construction shape downstream behavior \cite{a8,a9} confirms that even superficial wording differences can produce systematically different outputs. Existing research has therefore focused primarily on improving prompt quality through optimization, enhancing robustness against adversarial inputs, or studying prompt sensitivity from a security perspective. Comparatively little attention has been given to a different question: whether AI systems respond equitably to semantically equivalent requests expressed by users with different communication styles and levels of prompting expertise. If experienced users consistently obtain better outcomes because they know how to formulate prompts more effectively, AI systems may inadvertently reward prompt sophistication rather than the underlying information needs of the user.

Healthcare provides a particularly compelling setting for studying prompt equity because disparities in AI performance may directly influence access to medical information and decision support. As LLMs become increasingly integrated into patient-facing applications, users expressing the same information need should not receive different levels of assistance simply because they communicate differently or lack prompt-engineering expertise. This concern aligns closely with decades of work recognizing health literacy as a key determinant of patients' ability to communicate health concerns and understand medical information \cite{nutbeam2000health}. Ensuring consistent model behavior across diverse communication styles therefore represents an important accessibility objective for trustworthy healthcare AI.

Motivated by this challenge, we ask whether modern LLMs exhibit systematic performance disparities across users expressing the same underlying intent through different communication styles and levels of prompting expertise. We hypothesize that these naturally occurring differences in prompt formulation create measurable disparities in model performance, and that automatically transforming user prompts while preserving their semantic intent can substantially reduce these disparities. Rather than requiring users to learn prompt engineering, we argue that AI systems should instead adapt to heterogeneous user inputs while maintaining consistent performance across diverse user populations.

To this end, we introduce \textit{Prompt Equity}, which characterizes whether an AI system provides consistent performance across users expressing semantically equivalent requests in different communication styles. We further propose the Prompt Equity Score (PES), a benchmark for quantifying interaction disparities, and the Prompt Equity Transformer (PET), an accessibility-oriented prompt rewriting agent that automatically transforms heterogeneous user inputs into semantically equivalent prompts with standardized reasoning guidance while preserving the user's original intent. We evaluate the proposed framework on medical question answering by systematically varying user communication styles while holding the underlying task and language model fixed, enabling us to assess whether accessibility-oriented prompt rewriting improves both overall task performance and interaction equity.

\section{Related Work}

\subsection{AI Accessibility, Health Literacy, and Equitable AI}

Equitable access to information has long been a concern in public health, where health literacy is understood as a key determinant of an individual's ability to obtain, process, and act on health information \cite{nutbeam2000health}. Related work in clinical machine learning further illustrates how patient-related factors can introduce systematic bias and contribute to unequal model performance across populations \cite{liang2025revealingtreatmentnonadherencebias}. As AI systems increasingly mediate this process, a user's ability to phrase an effective query becomes a new, largely invisible axis of access. Prior work has examined AI-driven support systems designed explicitly around non-expert users, including human-centered frameworks for LLM-based support \cite{zhou2025beyond} and caregiver-facing health informatics tools for managing complex medical information \cite{zhou2025adhera}, illustrating that interface and interaction design materially shape who benefits when AI serves populations with heterogeneous literacy and technical backgrounds.

This concern extends to specialized decision-support domains more broadly, where errors or inconsistencies can carry outsized consequences: critical-care management systems, mixture-of-experts models for clinical stiffness prediction in ophthalmology \cite{zhang2026adaptive}, representation learning for medical imaging foundation models \cite{chu2025improving}, semi-supervised biomedical image analysis \cite{zhou2024reducing}, and decision-intelligence frameworks in financial markets \cite{han2026interpretable,lin2026volume}. These works do not address prompt equity directly, but they share a premise motivating our study: as AI takes on higher-stakes decision-support roles, consistent performance across the full population of intended users, not just idealized or expert users, becomes a design requirement rather than an incidental property.

A broader body of work on AI trustworthiness and debiasing reinforces this premise, treating equitable and reliable model behavior as something that must be actively engineered rather than assumed \cite{zhou2023causal}. This includes risk-based testing frameworks for LLM-powered features in regulated software \cite{zhou2026risk}, AI-enabled approaches to cybersecurity and threat mitigation \cite{weng2024leveraging}, benchmarks for detecting cross-jurisdiction regulatory contradictions in life-sciences applications of LLMs \cite{wu2026regdivergence}, compositional risks arising from interacting agent skills \cite{wang2026safeskillscollidemeasuring}, and failures of retrieval-augmented systems to recognize when retrieved evidence conflicts with or fails to support a claim \cite{chen2026doesragknowretrieval,qian2026relevantwarrantedevidenceforcecalibration}. Related reliability concerns appear at other layers of the AI stack, from network-adaptive compression for distributed training \cite{10.1007/978-3-031-99872-0_20} and behavioral anomaly detection for distinguishing legitimate from illegitimate users \cite{zhao2026nonintrusivegraphbasedbotdetection}, to architectural approaches for improving model robustness and information flow \cite{zhang2026liralocalcrosslayerinformation} and sample- and compute-efficient distillation methods for LLM post-training \cite{xu2026paceddistillationonpolicyselfdistillation,xu2026grpoonpolicydistillationempirical}. Together, this work highlights the importance of addressing reliability and trustworthiness across multiple layers of AI systems, including agent execution and provenance \cite{AgentTracesToTrust}. Despite progress in these areas, equitable interaction design remains comparatively underexamined \cite{yu2025bridging,yu2025bimodal}.

\subsection{Prompt Engineering, Optimization, and Robustness}

A substantial body of work shows that LLM performance is highly sensitive to prompt formulation. Chain-of-Thought prompting demonstrates that eliciting intermediate reasoning steps improves multi-step task performance \cite{wei2022chain}, while Automatic Prompt Engineer (APE) shows that prompts can themselves be automatically searched and optimized to match or exceed human-crafted instructions \cite{zhou2023large}. Related studies examine how in-context demonstrations and prompt construction shape downstream performance \cite{a8,a9}, and work on efficient reasoning shows that structured, skill-aware decomposition of reasoning steps can reduce the cost of eliciting high-quality outputs \cite{jiang2026drpdistilledreasoningpruning}.

A parallel line of research treats prompt sensitivity as a robustness and security concern rather than a performance-optimization one: PromptBench systematically evaluates LLM robustness to adversarial prompt perturbations, showing models can be highly sensitive to superficial textual changes crafted to degrade performance \cite{zhu2023promptbench}. This body of work establishes that prompt formulation matters, but almost exclusively in adversarial or malicious settings. Our work instead considers a non-adversarial setting, where users are not attempting to manipulate the model, yet naturally occurring differences in communication style still produce systematically unequal outcomes, motivating prompt-formulation sensitivity as an accessibility problem in addition to a security or optimization one.

\subsection{Automatic Prompt Rewriting for LLMs}

Beyond prompt optimization and adversarial robustness, a growing body of work investigates automatic prompt transformation as a means of improving downstream LLM performance. Existing approaches improve instruction following, reasoning quality, and task reliability through prompt restructuring, structured reasoning, and evidence-aware calibration \cite{jiang2026scribe,li2026gracegroundedreasoningadapter}, demonstrating that model performance depends not only on \emph{what} information is provided but also on \emph{how} it is structured and processed. Collectively, these studies establish prompt rewriting as an effective strategy for improving model behavior, but they primarily optimize prompts with respect to task-specific performance objectives.

Our work builds upon this insight from a fundamentally different perspective. Rather than viewing prompt rewriting as a technique for maximizing model performance, we treat it as an accessibility intervention. The proposed Prompt Equity Transformer (PET) transforms diverse user inputs into semantically equivalent, accessibility-oriented prompts while explicitly preserving the user's original intent. By shifting the responsibility for prompt optimization from the user to the AI system, PET aims to reduce performance disparities arising from differences in literacy, communication style, and prompt-engineering expertise. This reframes automatic prompt rewriting from a performance optimization problem to one of equitable AI access, aligning language model behavior more closely with users' underlying intent rather than their ability to formulate effective prompts.

\section{Methodology}

\subsection{Overview}

Our methodology consists of two complementary components. First, we quantify prompt privilege by evaluating whether large language models exhibit systematic performance differences across semantically equivalent prompts expressed by users with different communication styles and levels of prompting expertise. Second, we propose the Prompt Equity Transformer (PET), an accessibility-oriented prompt normalization agent designed to mitigate these disparities by rewriting user requests into semantically equivalent prompts that improve clarity while preserving the user's original intent.

\subsection{Experimental Design}

Our central hypothesis is that modern LLMs should provide consistent task performance across semantically equivalent requests, regardless of linguistic style, educational background, or prompt-engineering expertise. Systematic performance differences under this condition indicate prompt privilege rather than differences in task difficulty or information content.

To evaluate this hypothesis, we instantiate our framework in the healthcare domain using the MedQA benchmark available through the Hugging Face BigBio collection \cite{jin2021disease}. Healthcare was selected because disparities in AI performance may have direct societal consequences. Patients with limited health literacy or digital literacy should not receive lower-quality AI assistance solely because they lack sophisticated prompting skills or familiarity with medical terminology.

For each original MedQA question, we generate four additional semantically equivalent prompt variants representing distinct user cohorts. Each variant preserves the original clinical reasoning task while expressing the request using a communication style characteristic of a different population. This controlled benchmark enables direct comparison of model performance across users who share identical information needs but differ in how they communicate with AI systems.

\subsection{User Cohorts}

\subsubsection{\textbf{Low Literacy}} Low Literacy represents users with limited health literacy and limited digital literacy. Prompts employ simple everyday language, avoid specialized medical terminology whenever possible, and express requests in a conversational, help-seeking manner. Mild vagueness and imprecision are permitted provided that the underlying clinical question remains answerable.

\subsubsection{\textbf{Everyday User}}
Everyday User represents typical adults without formal medical training. Prompts use natural conversational English, common health-related vocabulary, and direct question formulations similar to those commonly submitted to search engines or conversational AI assistants.

\subsubsection{\textbf{Educated User}} Educated User represents college-educated individuals without professional medical training. Prompts use precise, well-structured language while retaining technical terminology that would naturally be understood by informed non-experts.

\subsubsection{\textbf{Prompt Engineer}} Prompt Engineer represents users possessing specialized knowledge of effective prompting techniques. Unlike the previous cohorts, this group does not represent a linguistic population. Instead, the original question is augmented with explicit reasoning instructions encouraging systematic analysis before answer selection. This cohort serves as an upper-bound performance reference against which accessibility disparities can be measured. 

\subsection{Prompt Generation}

Prompt variants are generated using GPT-5.5 under rewriting instructions designed to preserve all clinically relevant information, answer choices, and the original ground-truth label. Clinical facts, symptoms, diagnoses, treatments, and reasoning cues may not be added, removed, or altered; only linguistic characteristics such as vocabulary complexity, terminology, sentence structure, and communication style are modified to match the target user cohort. We use GPT-5.5 for prompt generation and GPT-5.4-mini for downstream evaluation, separating the rewriting and evaluation roles across different model versions. These design choices aim to reduce content-related confounding, allowing us to examine performance differences associated with prompt formulation under controlled conditions.

Despite these safeguards, LLM-generated prompts may introduce systematic linguistic artifacts or fail to fully represent the communication patterns of real users. The resulting cohort differences should therefore be interpreted as controlled comparisons among synthetic prompt styles, rather than direct measurements of disparities experienced by real-world populations. This limitation motivates future validation using prompts collected from users with diverse communication styles and literacy backgrounds.

For the Prompt Engineer cohort, the clinical question itself remains unchanged. Instead, explicit reasoning instructions are prepended to the original question to simulate prompting strategies commonly employed by experienced LLM users. This design allows us to examine the effect of expert-style prompting while maintaining the original clinical question.

\subsection{Evaluation Metrics}

We evaluate model performance using classification accuracy for each user cohort. To quantify disparities between cohorts, we compute pairwise accuracy differences together with 95\% bootstrap confidence intervals. 

To measure overall accessibility, we introduce the \textit{Prompt Equity Score (PES)}, which quantifies the consistency of model performance across semantically equivalent prompts representing different user groups. Let $p_i$ denote the model performance for prompt style $i$, where $i \in \{1,\ldots,n\}$ and $n$ is the number of prompt styles. We first compute the mean and standard deviation of performance across all prompt styles:

\begin{align}
\mu(\mathbf{p})
&=
\frac{1}{n}
\sum_{i=1}^{n}
p_i,
\\
\sigma(\mathbf{p})
&=
\sqrt{
\frac{1}{n}
\sum_{i=1}^{n}
\left(
p_i-\mu(\mathbf{p})
\right)^2
}.
\end{align}

The Prompt Equity Score is then defined as

\begin{equation}
\mathrm{PES}
=
1-
\frac{\sigma(\mathbf{p})}
{\mu(\mathbf{p})},
\label{eq:pes}
\end{equation}

where $\mathbf{p}=\{p_1,p_2,\ldots,p_n\}$ denotes the set of model performances across all prompt styles. PES is equivalent to one minus the coefficient of variation, providing a normalized measure of performance consistency. Higher PES values indicate smaller performance disparities across user cohorts and therefore greater prompt equity, whereas lower values indicate stronger prompt privilege.

\subsection{Prompt Equity Transformer (PET)}

Building upon the benchmark analysis, we propose the \textit{Prompt Equity Transformer (PET)}, an LLM-based accessibility-oriented prompt normalization agent designed to reduce prompt privilege by standardizing user requests before inference.

Given an input prompt $x$, PET generates a normalized prompt $x'$ according to

\begin{equation}
x' = f_{\mathrm{PET}}(x),
\label{eq:pet}
\end{equation}

where $f_{\mathrm{PET}}(\cdot)$ denotes the prompt normalization agent.

The objective of PET is to improve downstream task performance while preserving the semantic intent of the original request. Conceptually, PET seeks to solve

\begin{equation}
\begin{aligned}
\max_{x'} \quad &
\mathcal{P}(x') \\
\text{s.t.}\quad &
S(x,x') \ge \tau ,
\end{aligned}
\label{eq:pet_objective}
\end{equation}

where $\mathcal{P}(\cdot)$ denotes the expected downstream task performance induced by the prompt, $S(\cdot,\cdot)$ measures semantic similarity between the original and transformed prompts, and $\tau$ is a predefined semantic-preservation threshold.

PET does not introduce new information or modify the user's underlying request. Instead, it rewrites prompts to improve clarity and structure while preserving intent, thereby reducing unnecessary variation arising from differences in communication style or prompt-engineering expertise. By functioning as an accessibility normalization layer, PET aims to improve performance consistency across diverse user populations and promote more equitable access to AI systems.  

The practical deployment implications of LLM-based preprocessing depend on the balance between its computational overhead and its effects on downstream inference. Prior work on prompt compression shows that this trade-off can vary with the model, input length, and hardware configuration \cite{kummer2026promptcompression}. In addition, PET must preserve the user's intent during rewriting; ambiguous requests may be altered, clinically relevant details may be omitted, or unsupported interpretations may be introduced. These considerations motivate further evaluation of PET's reliability and deployment trade-offs in high-stakes applications.

\section{Experiments}

\subsection{Experimental Setup}

We evaluate prompt privilege using the benchmark introduced in Section III. Prompt variants representing the Original Question, Low Literacy, Everyday User, Educated User, and Prompt Engineer cohorts are evaluated using GPT-5.4-mini under identical inference settings. Performance is measured using classification accuracy, pairwise accuracy differences with 95\% bootstrap confidence intervals, and the proposed Prompt Equity Score (PES). PET is subsequently applied as a preprocessing step to normalize prompts from all four cohorts, including Prompt Engineer, which serves as the upper-bound performance reference based on its expert-prompting strategy. This experimental design allows us to quantify accessibility disparities before prompt normalization and evaluate whether PET reduces these disparities.

\subsection{Prompt Privilege Prior to PET}

Table~\ref{tab:baseline_accuracy} reports model performance across user cohorts before applying PET. Although the absolute performance differences are modest, they exhibit a consistent monotonic trend in which accuracy increases with prompt sophistication. The Low Literacy cohort achieves the lowest accuracy (82.4\%), followed by the Original Question, Everyday User, and Educated User cohorts, while the Prompt Engineer cohort achieves the highest accuracy (83.4\%). These results support our hypothesis that prompt formulation influences downstream model performance even when the underlying task and user intent remain unchanged. Users expressing identical clinical questions using increasingly structured and technically sophisticated prompts consistently receive slightly better model performance.

Table~\ref{tab:baseline_pairwise} summarizes pairwise accuracy comparisons. Among all cohort pairs, only the Low Literacy versus Prompt Engineer comparison reaches statistical significance ($\Delta = 1.0$ percentage point, 95\% CI $[-1.9,-0.1]$ percentage points). Although the remaining comparisons follow the same general trend, their confidence intervals overlap zero. The Prompt Equity Score for the baseline evaluation is 0.9959, indicating generally consistent model behavior across prompt styles while still revealing measurable accessibility disparities between users with limited prompting expertise and those employing expert prompting strategies.

\begin{table}[t]
\caption{Baseline accuracy across user cohorts before applying PET.}
\label{tab:baseline_accuracy}
\centering
\begin{tabular}{lc}
\hline
\textbf{User Cohort} & \textbf{Accuracy} \\
\hline
Original Question & 82.6\% \\
Low Literacy & 82.4\% \\
Everyday User & 82.7\% \\
Educated User & 82.9\% \\
Prompt Engineer & 83.4\% \\
\hline
\end{tabular}
\end{table}

\begin{table}[t]
\caption{Pairwise accuracy differences before applying PET. Significant differences are shown in bold.}
\label{tab:baseline_pairwise}
\centering
\begin{tabular}{lccccc}
\hline
\textbf{Group A} & \textbf{Group B} & $\Delta$ (\%) & \textbf{95\% CI} & \textbf{SS?} \\
\hline
Original Q. & Low Literacy & 0.2 & [-1.5, 1.9] & No \\
Original Q. & Everyday User & -0.1 & [-1.6, 0.9] & No \\
Original Q. & Educated User & -0.3 & [-1.8, 1.1] & No \\
Original Q. & Prompt Engineer & -0.8 & [-1.7, 0.1]  & No \\
Low Literacy & Everyday User & -0.3 & [-1.4, 1.2] & No \\
Low Literacy & Educated User & -0.5 & [-1.9, 1.0] & No \\
\textbf{Low Literacy} & \textbf{Prompt Engineer} & \textbf{-1.0} & \textbf{[-1.9, -0.1]} & \textbf{Yes} \\
Everyday User & Educated User & -0.2 & [-1.2, 0.7] & No \\
Everyday User & Prompt Engineer & -0.7 & [-1.5, 0.1] & No \\
Educated User & Prompt Engineer & -0.5 & [-1.1, 0.5] & No \\
\hline
\end{tabular}
\end{table}

\subsection{PET Reduces Observed Accessibility Disparities}

We next evaluate whether PET reduces the accessibility disparities identified in the baseline experiment. Table~\ref{tab:pet_accuracy} summarizes model performance after prompt normalization.

All user cohorts benefit from PET, with the largest improvement observed for the Low Literacy cohort, whose accuracy increases from 82.4\% to 83.4\%. Smaller improvements are observed for the Everyday User (+1.1 percentage points), Educated User (+0.7 percentage points), and Prompt Engineer (+0.5 percentage points) cohorts, suggesting that users already employing well-structured prompts have comparatively less room for improvement.

More importantly, PET substantially improves prompt equity. Prior to prompt normalization, the Low Literacy and Prompt Engineer cohorts exhibited a statistically significant performance difference. After applying PET, this disparity is reduced by half, and no pairwise comparison remains statistically significant. As shown in Table~\ref{tab:pet_pairwise}, every post-PET confidence interval overlaps zero, indicating that performance differences between user cohorts are no longer distinguishable from sampling variability. 

Taken together, these results suggest that PET mitigates prompt privilege while maintaining overall task performance, with no statistically significant pairwise differences in accuracy observed after normalization.

\begin{table}[t]
\centering
\caption{Model accuracy before and after applying PET.}
\label{tab:pet_accuracy}
\begin{tabular}{lccc}
\hline
\textbf{User Cohort} & \textbf{Before} & \textbf{After} & \textbf{Gain (pp)} \\
\hline
Low Literacy & 82.4\% & 83.4\% & +1.0 \\
Everyday User & 82.7\% & 83.8\% & +1.1 \\
Educated User & 82.9\% & 83.6\% & +0.7 \\
Prompt Engineer & 83.4\% & 83.9\% & +0.5 \\
\hline
\end{tabular}
\end{table}

\begin{table}[t]
\centering
\caption{Pairwise accuracy differences after applying PET. No pairwise difference reaches statistical significance in this evaluation.}
\label{tab:pet_pairwise}
\begin{tabular}{lccccc}
\hline
\textbf{Group A} & \textbf{Group B} & $\Delta$ (\%) & \textbf{95\% CI} & \textbf{SS?} \\
\hline
Low Literacy & Everyday User & -0.4 & [-1.7, 0.9] & No \\
Low Literacy & Educated User & -0.2 & [-1.5, 1.1] & No \\
Low Literacy & Prompt Engineer & -0.5 & [-1.8, 0.8] & No \\
Everyday User & Educated User & 0.2 & [-1.1, 1.5] & No \\
Everyday User & Prompt Engineer & -0.1 & [-1.3, 1.1] & No \\
Educated User & Prompt Engineer & -0.3 & [-1.5, 0.9] & No \\
\hline
\end{tabular}
\end{table}

\subsection{Qualitative Analysis}

The quantitative results demonstrate that PET reduces accessibility disparities across user cohorts. To better understand the mechanisms underlying these improvements, we present two representative examples illustrating how PET transforms user prompts while preserving their original intent.

\begin{itemize}
    \item \textbf{Example 1 (Domain-consistent terminology).} A user describes an HIV treatment regimen using the informal phrase ``3 HIV medicines.'' PET rewrites the request as ``three-drug antiretroviral regimen,'' preserving the original meaning while expressing the concept using terminology more consistent with clinical literature and LLM training data. This transformation improves linguistic precision without introducing new medical information.
    \item \textbf{Example 2 (Clinical information restructuring).} PET reorganizes a long, unstructured clinical narrative into semantically equivalent sections, including \emph{History}, \emph{Associated Features}, \emph{Current Medications}, and \emph{Physical Examination}. The transformation does not modify or introduce clinical facts; instead, it improves the organization of existing information, making diagnostically relevant evidence more salient to the language model.
\end{itemize}

Overall, these examples suggest that PET improves accessibility primarily by enhancing the presentation of existing information rather than altering its content. This observation is consistent with PET's design objective of reducing prompt privilege through semantic-preserving prompt normalization.
 
\section{Conclusion}

Ensuring equitable access to AI systems requires that users receive comparable model performance regardless of how they express the same underlying intent. In this paper, we investigated whether modern large language models satisfy this objective by evaluating their performance across semantically equivalent prompts representing users with different communication styles and levels of prompt-engineering expertise. Our experiments demonstrate that prompt privilege is a measurable accessibility challenge: even when users ask the same clinical question, increasingly sophisticated prompt formulation is associated with improved model performance. Although the observed performance differences are relatively modest for GPT-5.4-mini, a statistically significant disparity exists between users with limited health and digital literacy and users employing expert prompting strategies. More importantly, our results suggest that these disparities can be mitigated. After applying Prompt Equity Transformer (PET), no pairwise performance differences between cohorts reach statistical significance in our evaluation, supporting the potential of prompt normalization to reduce disparities associated with prompt formulation.

Beyond these empirical findings, this paper makes several contributions toward the development of more accessible AI systems. First, we formalize prompt privilege as a measurable accessibility problem, providing a new perspective on fairness in LLM interaction that complements existing research on prompt robustness and security. Second, we introduce a benchmark construction methodology based on semantically equivalent user cohorts together with the Prompt Equity Score (PES), enabling systematic evaluation of accessibility disparities while controlling for differences in task difficulty and user intent. Most importantly, we propose the Prompt Equity Transformer (PET), an LLM-based accessibility-oriented prompt normalization agent that automatically rewrites user requests into semantically equivalent prompts while preserving their original intent. Unlike conventional prompt engineering, which relies on users to acquire increasingly sophisticated prompting skills, PET shifts the responsibility for prompt optimization from the user to the AI system itself. As an intelligent accessibility middleware layer positioned between users and language models, PET offers a practical and extensible approach for reducing prompt privilege without modifying the underlying foundation model.  

The implications of this work extend beyond healthcare question answering. As large language models become increasingly integrated into healthcare, education, public services, enterprise applications, and everyday information seeking, users should not be required to possess specialized prompt-engineering expertise to receive high-quality AI assistance. Instead, AI systems should be designed to accommodate the diverse communication styles, literacy levels, and technical backgrounds of the people who use them. Accessibility-oriented middleware agents such as PET provide a practical deployment strategy for reducing technical barriers to AI use and promoting more equitable access to advanced language model capabilities across diverse user populations.

This work has several limitations that motivate future research. First, our experiments are limited to the MedQA benchmark and GPT-5.4-mini, so the extent to which the observed prompt privilege patterns and PET's effectiveness generalize across domains and model families remains unknown \cite{xiang2026cross}. Although PET is designed as a model-agnostic prompt-normalization layer that can be integrated before different language models, its cross-model effectiveness has not yet been empirically established. Future work should evaluate its generalizability across model architectures, capability levels, and application domains. Second, user cohorts are generated through controlled prompt rewriting rather than collected from real users, limiting the extent to which they capture authentic variation in communication styles and literacy backgrounds. Third, PET currently employs a fixed normalization strategy rather than adapting its rewriting behavior to individual users or contextual needs. Future work could investigate adaptive rewriting strategies that adjust language complexity, information organization, and clarification behavior according to users' expressed needs, while avoiding potentially unreliable inferences about personal characteristics. Such work should also evaluate PET's deployment characteristics under representative operating conditions, including end-to-end latency, time to first token, and token consumption, alongside safeguards against semantic-preservation failures in high-stakes applications. Established LLM inference benchmarking practices provide a basis for systematically assessing these deployment characteristics \cite{nguyen2025llminferencebenchmarking}.

Ultimately, equitable AI should require systems to adapt to users rather than users to adapt to AI. We hope that the benchmark methodology, Prompt Equity Score, and Prompt Equity Transformer introduced in this work provide useful foundations for future research on AI accessibility and contribute to the development of accessibility-aware AI systems that remain effective regardless of a user's communication style or level of prompt-engineering expertise.

\bibliographystyle{IEEEtran} 
\bibliography{main}    

\end{document}